# A PROBABILISTIC MODEL FOR SENSOR VALIDATION

**P.H. Ibargüengoytia**
Instituto de Investigaciones
Eléctricas, A.P. 1-475
Cuernavaca, Mor., 62001, México
pibar@iie.org.mx

**L.E. Sucar**
Instituto Tecnológico y de
Estudios Superiores de Monterrey
Campus Morelos, A.P. 99-C
Cuernavaca, Mor., 62050, México
esucar@campus.mor.itesm.mx

**S. Vadera**
University of Salford
Dept. of Mathematics
and Computer Science
Salford, M5 4WT, U.K.
S.Vadera@mcs.salford.ac.uk

## Abstract

The validation of data from sensors has become an important issue in the operation and control of modern industrial plants. One approach is to use knowledge based techniques to detect inconsistencies in measured data. This article presents a probabilistic model for the detection of such inconsistencies. Based on probability propagation, this method is able to find the existence of a possible fault among the set of sensors. That is, if an error exists, many sensors present an *apparent* fault due to the propagation from the sensor(s) with a real fault. So the fault detection mechanism can only tell if a sensor has a *potential* fault, but it can not tell if the fault is real or apparent. So the central problem is to develop a theory, and then an algorithm, for distinguishing real and apparent faults, given that one or more sensors can fail at the same time. This article then, presents an approach based on two levels: (i) probabilistic reasoning, to detect a potential fault, and (ii) constraint management, to distinguish the real fault from the apparent ones. The proposed approach is exemplified by applying it to a power plant model.

## 1 INTRODUCTION

Computing is playing an increasingly important role in domains like communications, medicine, and industry. Examples of industrial applications include the control of advanced manufacturing plants, power generation, power distribution, and chemical processes. These applications require the utilization of several methodologies that have emerged from the area of artificial intelligence (AI). In general, AI methods are moving towards more realistic domains that require cooperation between several fields of research. This paper describes an ongoing research project in the utilization of AI methods to solve the problem of sensor validation. Although the techniques presented here can be considered as general, the specific application is in the power plants domain.

The approach proposed in this paper has two layers:

- a prediction layer: which is used to predict the expected values of the sensors and identify *potential* faults;

- a constraint satisfaction layer: which is used to distinguish the faulty sensor(s) from the apparently faulty ones.

Both layers make use of a probabilistic network model. A probabilistic or Bayesian network [Pearl, 1988] is a directed acyclic graph (DAG) whose structure corresponds to the dependency relations of the set of variables represented in the network (nodes), and which is parameterized by the conditional probabilities (links) required to specify the underlying distribution. In this case, the nodes correspond to the sensors that constitute the model. The structure of the network makes explicit the dependence and independence relations between the variables.

In this approach, with the use of probability propagation, a prediction is made of a variable's value based on other parameters. If this predicated value deviates from the actual value given by a sensor, by some predefined margin, then some fault can be assumed. But the fault detection mechanism can only tell if a sensor has a *potential* fault, but it can not tell if the fault is real or apparent. The central problem is to develop a theory, and then an algorithm, for distinguishing real and apparent faults, considering that one or more sensors can fail at the same time. For this, the structure of the model is considered, which produces a set of constraints that has to be solved to determine the faulty sensor(s). This article then, presents an approach based in two levels: (i) probability propagation, to detect a potential fault, and (ii) constraint management, to distinguish the real faulty from the apparent ones.

The paper is organized as follows. Section 2 introduces the problem and summarizes previous approaches. Section 3 presents the approach with the aid of a simple example. Section 4 presents the ideas formally. Section 5 describes a real example that shows the complete technique. Finally, section 6 presents the conclusions and future work.

## 2 SENSOR VALIDATION

The validation of data from sensors has become an important issue in the operation and control of modern industrial plants. Usually, the control system can not detect significant deviations from the expected values given the design working point, for example of the gas turbine in a power plant. Conversely, an experienced operator is capable of detecting such deviations of a variable by direct observation of the related variables and consequently, avoids false plant trips. This project proposes the modelling of the operator's experience in the detection of sensor failures.

Typical solutions to this problem, particularly in critical systems where security is essential, include the use of:

- *Hardware redundancy and majority voting:* where hardware is duplicated and a voting algorithm is used to exclude faulty sensors. This is possible in applications such as civilian aircraft or part of the nuclear industry [Yung and Clarke, 1989]. However, for many industrial plants, these techniques are not feasible where, for example, adding further sensors might weaken the walls of the pressure vessels.

- *Analytical redundancy:* in which all process, actuators and sensors are monitored centrally. Examples of these techniques are *generalized likelihood ratio* (GLR) [Willsky and Jones, 1976], and *failure sensitive filters* [Massoumnia, 1986].

However, these approaches can require the development of mathematical or knowledge based models whose solution require expensive computer power. Additionally, they are very expensive and demand an enormous amount of expertise to use them in a different process or even make a modification of the monitored system. Modern techniques, from where this project is motivated, include a decentralised and hierarchical approach [Yung and Clarke, 1989]. A survey of some of these techniques can be found in [Basseville, 1988].

Previous stages in the development of this project included some experiments in the validation of signals in power plants [Ibargüengoytia et al., 1995]. These experiments were based on the following assumption: each sensor is validated independently, i.e., each variable was considered as the hypothesis while some other variables were considered as correct evidence. However, a real solution of the problem requires a different set of assumptions to be taken. For example, if the turbine velocity is validated utilizing only the signals of temperature and pressure, and if the reasoning reports a faulty sensor, it is impossible to define which sensor was the faulty one. In this example, if the temperature sensor fails and it is utilized to detect a fault in the velocity, the system will certainly report a failure on the velocity reading. This could be a wrong conclusion.

Such an approach, of course, requires the help of domain experts to identify the dependencies of the variables and must also take into account of the following characteristics:

- The sensors can provide erroneous information.
- Information is available all the time, i.e., all sensors can be observed as evidence or considered as an hypothesis at any time.
- The system must respond within a real time environment.
- The application considers the possibility of multiple faults.

## 3 THE APPROACH PROPOSED

This section presents the approach proposed through a very simple example. Assume the model of the gas turbine in a power plant shown in Fig. 1[1]. The root node $m$ represents the reading of the Megawatts generated in the plant. The temperature is represented by node $t$ and the pressure by $p$. Finally, $g$ represents the fuel supplied to the combustion chamber. The validation process starts assuming that the sensors, one by one, are suspect. By probabilistic reasoning, the system decides if the reading of the sensor is correct based on the values of the most related variables. This process is carried out for each one of the variables that is required to be validated. The most closely related variables for each sensor consist of the *Markov blanket* of the sensor variable. A Markov blanket is defined as the set of variables that make a variable independent from the others. In a Bayesian network, the following three sets of neighbours is sufficient for forming a Markov blanket of a node: the set of direct predecessors, direct successors, and the direct predecessors of the successors (i.e., parents, children, and spouses). The set of variables that constitutes the Markov blanket of a variable can be seen as a protection of this variable against changes of variables outside the blanket. This means that, in order to analyze a variable, it is only needed to know the value of all variables in its blanket. For example, in Fig. 1 a Markov blanket of $t$ is $\{m, g\}$, while a blanket of $g$ consists of $\{t\}$ only. Table 1 shows the Markov blankets of each one of the variables in the model of Fig. 1. Considering these blankets, probabilistic reasoning is performed utilizing the reduced models for each variable as shown in Fig. 2. In (a), the equivalent model of $m$ is shown where the absence of $g$ indicates that this variable is out of $m$'s Markov blanket. In (b), the model to predict $t$ indicates that the changes of $p$ are not considered given a value of $m$. The same for $p$ and $g$ in (c) and (d).

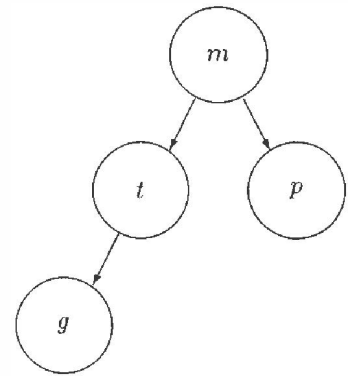

Figure 1: Simple tree representing the turbine generation model.

---
[1]This is a simplified model of the gas turbine. The directions of the arcs do not imply causality.

Table 1: Markov blankets of the turbine model.

| process variable | Markov blanket |
|---|---|
| $m$ | $\{t, p\}$ |
| $t$ | $\{m, g\}$ |
| $p$ | $\{m\}$ |
| $g$ | $\{t\}$ |

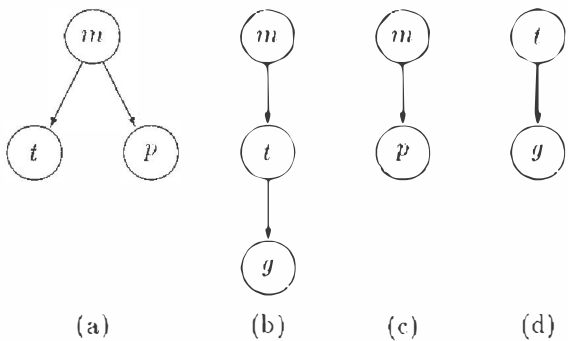

Figure 2: Equivalent models for the variables. (a) for $m$, (b) for $t$, (c) for $p$, and (d) for $g$.

Assume that the temperature sensor suffers catastrophic damage, e.g., the wires were cut. Starting the validation process with $m$, since $t$ participates in the validation [see Fig. 2(a)], and due to the failure in $t$, the reasoning will indicate that there is a failure in $m$. Next, the validation for $t$ will of course indicate the existence of a failure. The validation of $p$ will indicate that this sensor is working properly. Finally, the validation of $g$, given its equivalent model shown in Fig. 2(d), will also indicate a failure in the sensor. So, even after the probabilistic reasoning, there is still confusion: which are real and which are apparent faults?

The use of a constraint satisfaction system is required. In this case, the presence of a faulty sensor causes a constrained area of manifestation which forms a context. The contexts can be arranged in a lattice as shown in Fig. 3. The lower node represents the no fault context of the system. The upper layers represent an incremental assumption of faulty sensors. The top node represents a context where all the sensors are reported faulty.

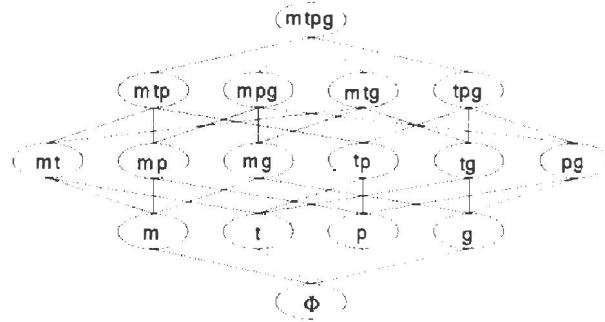

Figure 3: Lattice with the four variables, $m, t, p$, and $g$, for the model in Fig. 1.

Every step in the probabilistic reasoning generates a constraint for the final detection of the sensor in fail. Starting at the bottom of the lattice of Fig. 3, each step will make a transition between the nodes of the lattice. Figure 4 shows the transitions from the bottom node ($\phi$) to the final node: $\{m, t, g\}$.

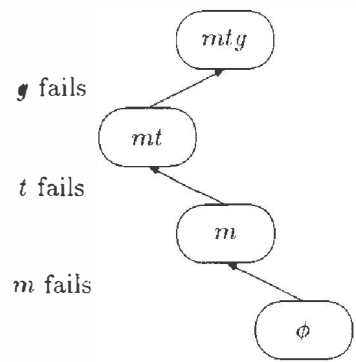

Figure 4: Trajectory followed in the lattice by the probabilistic reasoner.

This paper shows that there exists a correspondence between a fault in a sensor and the node of the lattice obtained when propagating the failures reported by the probabilistic reasoning. In this example, when $t$ fails, the corresponding lattice node is $\{m, t, g\}$. This node corresponds to the set of variables that forms the Markov blanket of a variable plus the variable itself.

Table 2 shows the Markov blankets and the lattice's node corresponding to the constraints propagation for each variable in fault. Then, if a propagation on the lattice finishes in $\{m,t,p\}$ it signifies that the fault is in sensor $m$. Lattice node $\{m,p\}$ corresponds to $p$, and finally, $\{t,g\}$ corresponds to $g$.

Table 2: Markov blankets of the simple turbine model.

| process variable | Markov blanket | lattice node |
|---|---|---|
| $m$ | $\{t,p\}$ | $\{m,t,p\}$ |
| $t$ | $\{m,g\}$ | $\{m,t,g\}$ |
| $p$ | $\{m\}$ | $\{m,p\}$ |
| $g$ | $\{t\}$ | $\{t,g\}$ |

With this mechanism, even if there exist many apparent faults, the propagation on the lattice distinguishes which sensor contains the real fault. Table 2 considers only single faults. However, the method assures that if the lattice propagation finishes in a node not included in Table 2, it signifies that there exists more than one sensor with a fault. Section 4 explains and demonstrates these mechanisms.

## 4 A THEORY FOR SENSOR VALIDATION

The probabilistic model for sensor validation consists of a Bayesian network as defined by Pearl [Pearl, 1988]. That is, a directed acyclic graph $G$ which is a *minimal I-map* of the dependency model $M$ for a probability distribution $P$. $G$ is an *I-map* or *independency map* of $M$ if

$$I(X,Z,Y)_M \Longleftarrow <X\mid Z\mid Y>_G \qquad (1)$$

where $I(X,Z,Y)$, $X,Y,Z$ are subsets of $V$, denotes conditional independence of $X$ and $Y$ given $Z$, and $<X\mid Z\mid Y>_G$ represents a graph $G$ where the subset $Z$ of nodes intercepts all paths between the nodes of $X$ and those of $Y$. $G$ is an *I-map* of $M$ if every conditional independence condition (according to the *D-separation* criteria) displayed in $G$ corresponds to a valid conditional independence relationship in $M$. It is a *minimal I-map* if none of its links can be deleted without destroying its *I-mapness* [Pearl, 1988].

**Definition 1:** Given a probability distribution $P$ on a set of variables $V$, a DAG $G=(V,E)$ ($V$ is a set of nodes, $E$ is a set of links) is a *Bayesian Network* iff $G$ is a *minimal I-map* of $P$.

A *Markov blanket* for any node $X$ in a Bayesian network is a subset of $V$ which makes it independent from the other variables.

**Definition 2:** A *Markov blanket* $MB(X)$ of any variable $X \in V$ is a subset $S \subset V$ where $X \notin S$ for which

$$I(X,S,V-S-X) \qquad (2)$$

In a Bayesian network, the *Markov blanket* of a node $X$ can be formed by the union of its direct parents $PA(X)$, its direct successors $SU(X)$, and all direct parents of the latter $SP(X)$ ($X$'s spouses). This follows from the axioms of conditional independence and the *Soundness* theorem [Pearl et al., 1990]. Although there may be other Markov blankets, only this type of blankets are considered.

In using a Bayesian network representation for sensor validation, the following assumptions are made:

1. Observability: all the variables (sensors) can be measured directly [2].

2. Fault detection: if there is an error in sensor $X$ it can always be detected. That is $x_o \neq x_p$, where $x_o$ is the observed value, and $x_p$ the *predicted value*. The predicted value is the one with highest probability obtained by probability propagation from its neighbours, $MB(X)$, assuming $X$ is unknown. This is called a *real fault* denoted by $Fr(X)$.

3. Fault propagation: if a sensor $Y$ has a real fault $Fr(Y)$, and $Y \in MB(X)$, a fault in $X$ will be detected, that is $x_o \neq x_p$. This is called an *apparent fault* denoted by $Fa(X)$. In general, $Y$ could be a set of sensors such that $Y \subset G$.

If a sensor (variable) $X$ has a real fault and/or apparent fault then it is called a *potential fault* $Fp(X)$. The fault detection mechanism can only tell if a sensor has a potential fault, but (without considering other sensors) it can not tell if the fault is real or apparent. So the central problem is to develop a theory, and then an algorithm, for distinguishing real and apparent faults, considering that one or more sensors can fail at the same time.

**Lemma 1 (symmetry):** Let $X$ be a node in a Bayesian network $G$ with a Markov blanket[3] $MB(X)$, $X \in MB(Y_i)$ iff $Y_i \in MB(X)$, $\forall Y_i \in G$. That is, $X$ is in the Markov blanket of all the variables that are in $MB(X)$, and it is only in these Markov blankets.

**Proof:** First, the proof that if $Y \in MB(X)$ then $X \in MB(Y)$. Given that $MB(X) = PA(X) \cup SU(X) \cup SP(X)$, then $Y \in PA(X)$ or $Y \in SU(X)$ or $Y \in SP(X)$, so $X \in SU(Y)$ or $X \in PA(Y)$ or $X \in SP(Y)$, respectively. In any case, $X \in MB(Y)$. Next, the proof that if $Y \notin MB(X)$ then $X \notin MB(Y)$. By Definition 2, $I(X,MB(X),G-MB(X)-X)$, and by the *Symmetry* and *Decomposition* axioms [Pearl et al., 1990] $I(Y_i,MB(X),X)$, $\forall Y_i \in G-MB(X)-X$. Thus $X$ is not in $MB(Y)$.□

---

[2] A one to one correspondence between nodes, variables, and sensors is considered.

[3] This lemma and the subsequent theorems apply to Markov blankets formed by the direct parents, direct successors, and direct parents of the latter.

The *extended Markov blanket* $EMB(X)$ is defined as the union between the Markov blanket of a variable and the variable, i.e., $EMB(X) = X \cup MB(X)$.

**Theorem 1:** If there is an error in sensor $X$, it will produce a *potential fault* in $X$, and all the sensors in $MB(X)$, and no other sensor.

**Proof:** From assumption 2, an error in $X$ produces a potential fault in $X$. From Lemma 1, $X$ is an element of the MB of all sensors $Y_i \in MB(X)$, so by assumption 3, an error in $X$ produces potential faults in all sensors in $MB(X)$. Finally, from Lemma 1 $X$ is not an element of any other MB, so no other potential fault will be detected (assuming $X$ is the only sensor in fault).□

**Corollary 1:** Assuming a single (only one sensor fails) real fault, a *potential fault* in all the sensors in $S \subset G$ implies a *real fault* in a sensor $X$ such that $EMB(X) = S$, and $X \in S$.

Given that an error in a sensor produces a potential fault in all the sensors in its EMB and only in those, a potential fault in all sensors in $S$ implies that a sensor in fault has its $EMB = S$ (assuming one real fault).

**Corollary 2:** If all Markov blankets are different in $G$, $MB(Y) \neq MB(Z), \forall Y, Z \in G, Y \neq Z$, then all single *real faults* $Fr(X)$ can be distinguished in $G$. In this case only the nodes in $EMB(X)$ will have a *potential fault*.

The validity of Corollary 2 follows from Corollary 1 and the fact that all MB are different.

The problem is that two or more variables could have the same Markov blanket. Such is the case of *leaf* nodes with the same parent in a tree.

**Corollary 3:** If there is an error in sensor $X$ with $EMB(X)$, and also an error in $Y$ with $EMB(Y)$, they will produce *potential faults* in all nodes $Z \in EMB(X) \cup EMB(Y)$. In general, if there are errors in sensors $X_i, i = 1, ..., m$, they will produce potential faults in all nodes $Z \in EMB(X_1) \cup ... \cup EMB(X_m)$

Corollary 3 follows directly from Theorem 1, assuming that two or more errors will not *cancel* each other (i.e., if $Z \in MB(X)$ and $Z \in MB(Y)$ and both, $X$ and $Y$ fail, still a potential fault will be detected in $Z$).

**Theorem 2:** If there is an error in sensor $X$ with $EMB(X)$, and $Y \in EMB(X)$ with $EMB(Y) \subset EMB(X)$, and multiple faults (more than one sensor can fail simultaneously) are considered, then there is no distinction between $Fr(X)$ or $Fr(X) \wedge Fr(Y)$.

**Proof:** By Theorem 1, $Fr(X)$ will produce apparent faults in $EMB(X)$. $Fr(X) \wedge Fr(Y)$ will produce apparent faults in $EMB(X) \cup EMB(Y)$ by Corollary 3. So if $EMB(Y) \subset EMB(X)$, then $EMB(X) \cup EMB(Y) = EMB(X)$ so both cases are indistinguishable.□

**Theorem 3:** Multiple faults can be distinguished if all the *extended Markov blankets* of the sensors with errors are disjoint.

**Proof:** It follows from Corollary 2 and 3.□

**Theorem 4:** If the set of nodes $S$ with apparent faults in $G$ is different from all $EMB(X_i), \forall X_i \in G$, there must be multiple (at least 2) *real faults* in $G$. The sensors $X_i$ such $EMB(X_i) \subset S$ can have real faults, an only these sensors.

**Proof:** From Theorem 1, a real fault in $X$ produces apparent faults in and only in the set of sensors in $EMB(X)$. So a single fault can not produce a set $S$ of potential faults different from all EMB in $G$. From Corollary 3, the sensors whose EMB is a subset of $S$ can be in fault, and by Theorem 1, only these sensors.□

Based on the theory described above, an algorithm is required so that, once the model has been established, and the Markov and extended Markov blankets have been defined, the detection of real failures can be carried out. The proposed algorithm for sensor validation is the following:

1. Obtain the model (i.e., the Bayesian network) of the application process.

2. Make a list of the variables to be validated and build a table of EMB for each variable.

3. Take each one of the variables to be checked as the hypothesis, instantiate the variables that form the Markov blanket of the hypothesis, and propagate the probabilities to obtain the posterior probability of the variable given the evidence.

4. Compare the predicted value (the posterior probability) with the current value of the variable and decide if an error exists.

5. Repeat steps 3 and 4 until all the variables in the list have been examined and the set of sensors with apparent faults ($S$) is obtained.

6. Compare the set of sensors in fault obtained in step 5, with the table of the EMB for each variable:

    (a) If $S = \phi$ there are no faults.
    (b) If $S$ is equal to the EMB of a variable $X$, and there is no other EMB which is a subset of $S$, then there is a single real fault in $X$.
    (c) If $S$ is equal to the EMB of a variable $X$, and there are one or more EMB which are subsets of $S$, then there is a real fault in $X$, and possibly, real faults in the sensors whose EMB are subsets of $S$.
    (d) If $S$ is equal to the union of several EMB and all these are disjoint, there are multiple real faults in all the sensors whose EMB are in $S$.
    (e) If none of the above cases is satisfied, then there are multiple faults but they can not be distinguished. All the sensors whose EMB are subsets of $S$ could have a real fault.

Notice that the propagation on the lattice is an indexing of a table, i.e., no calculations are required. This is a very important feature for a system running in real time.

The next section explains the algorithm proposed in a real example, taken from a thermoelectrical power plant.

## 5  SENSOR VALIDATION IN A POWER PLANT

In order to demonstrate the ideas contained in this article, a module of a combined cycle power plant was chosen: the gas turbine. Figure 5 shows a simplified schematic diagram of the type of gas turbines at the *Dos Bocas* and *Gomez Palacio* power plants in Mexico.

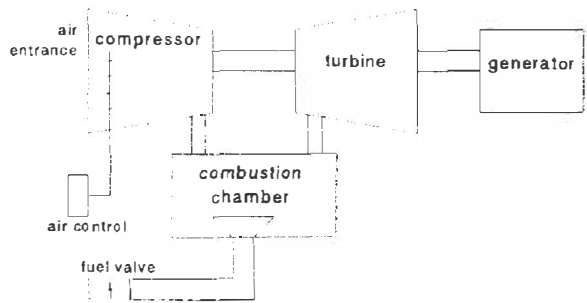

Figure 5: Simplified schematic diagram of a gas turbine.

A gas turbine consists fundamentally of four main parts: the compressor, the combustion chamber, the turbine itself and the generator. The compressor feeds air to the combustion chamber, where the gas is also fed. Here, the combustion produces high pressure gases at high temperature. The expansion of these gases in the turbine produces the turbine rotation with a torque that is transmitted to the generator in order to produce the electric power output. The air is regulated by means of the *inlet guide vanes* (IGV) of the compressor, and a control valve does the same for the gas fuel in the combustion chamber. The control valve is commanded by the control system or by the operator in the manual operation mode, and its aperture can be read by a position sensor. The temperature at the blade path, which is the most critical variable, is taken along the circumference of the turbine.

Among all variables that participate in the gas turbine, only a few are directly measured by the sensors. Since the blade path temperature is the most critical variable, it is obtained through sixteen thermocouple sensors located all around the turbine. From these sixteen, three sets of averages are taken by analog circuitry. These values are then averaged in order to obtain a single value for the temperature. The operator is informed of the average temperature, which is also used by a control strategy to protect the engine. Table 3 shows the list of variables of the model of Fig. 6 including the explanation of the process variable, their name, and their corresponding Markov blankets. From Table 3, it is easy to see that the lat-tice nodes for variables $t1$, $t2$, and $t3$ is exactly the same, i.e., $\{T, t1, t2, t3, f1, f2\}$. Hence according to section 4, failures amongst these sensors can not be distinguished by the system.

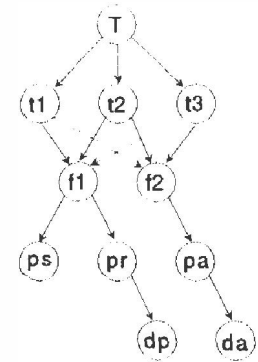

Figure 6: Model of the gas turbine example.

Table 3: Variables in the example.

| process variable | | Markov blanket |
|---|---|---|
| Selected blade path temp. | $T$ | $\{t1, t2, t3\}$ |
| Blade path temp. avg. 1 | $t1$ | $\{T, t2, t3, f1, f2\}$ |
| Blade path temp. avg. 2 | $t2$ | $\{T, t1, t3, f1, f2\}$ |
| Blade path temp. avg. 3 | $t3$ | $\{T, t1, t2, f1, f2\}$ |
| Flow of gas | $f1$ | $\{t1, t2, t3, ps, pr\}$ |
| Flow of air | $f2$ | $\{t1, t2, t3, pa\}$ |
| Gas fuel pressure supply | $ps$ | $\{f1\}$ |
| Real fuel valve position | $pr$ | $\{f1, dp\}$ |
| Real IGV position | $pa$ | $\{f2, da\}$ |
| Position demand fuel valve | $dp$ | $\{pr\}$ |
| Position demand IGV's | $da$ | $\{pa\}$ |

In this example, if the node $\{t1, t2, t3, ps, pr\}$ is obtained in the propagation on the lattice, according to Table 3, the fault is undoubtedly in the sensor measuring the variable $f1$ or flow of gas. Contrary to the example of Fig 1 which is very simple, this example allows the case of multiple faults also to be shown. Suppose that the sensor of the real fuel valve position ($pr$) fails together with the sensor of the position demand of IGVs ($da$). The lattice node of $pr$ is $\{f1, pr, dp\}$, and the $da$ node is $\{pa, da\}$. Consequently, the lattice node of the combined failure of $pr$ and $da$ is the union of their corresponding lattice nodes, i.e., $\{f1, pr, pa, dp, da\}$. Finally, if there exists a fault in sensor $dp$ and sensor $pr$, the resulting union between both extended Markov blankets is given by $\{f1, dp, pr\}$ which is exactly the same as the lattice node of $pr$. In this case, the model can only ensure that there exists a fault in $pr$ but it can not distinguish the double fault in $pr$ and $dp$.

# 6  CONCLUSIONS AND FUTURE WORK

This paper has presented an approach to detecting inconsistencies in the readings of sensors in industrial plants. This approach, based on Bayesian networks and constraint satisfaction, possesses also the advantage that much of the processing is performed off line, i.e., before the system operates in the plant. This characteristic will help in the real time performance required in most of the industrial applications. With the use of probability propagation, a prediction is made of a variable's value based on other parameters. If this predicated value deviates from the actual value given by a sensor, by some predefined margin, then some fault can be assumed. This fault detection mechanism can only tell if a sensor has a *potential* fault, but it can not tell if the fault is real or apparent. A theory and algorithm were developed for distinguishing real and apparent faults, considering that one or more sensors can fail at the same time. For this, the structure of the model is considered, which produces a set of constraints that has to be solved to determine the faulty sensor(s). The approach is based on two levels: (i) probability propagation, to detect a potential fault, and (ii) constraint management, to distinguish the real faults from the apparent ones. The method was applied to a simplified model of a gas turbine.

The main limitation of the proposed algorithm is that in some cases, it is not possible to identify precisely the real fault among all the sensors with potential faults. The cases when no exact answer is provided are summarized as follows:

- two or more sensors with the same EMB,
- a double fault where one EMB is a subset of the other,
- multiple faults in which some of the EMB fall in the previous cases.

For example, in Fig. 6, variables $t1$, $t2$, and $t3$ fall en the first case. In Fig. 1 there is an example of the second case. The EMB of $p$ is $\{m, p\}$ and the $m$'s EMB is $\{m, p, t\}$. Here, if $m$ and $p$ fail at the same time, this mechanism can only inform that $m$ failed but it can not tell if $p$ failed or not.

In order to be applied in a real plant, the technique must address additional issues.

First, the techniques described in this paper work well when the sensor fails catastrophically. However, a real problem faced by the operators in power plants is *slow* failures such as decalibration. Different probabilistic mechanisms have to be included in order to detect such slow deviations, for example, *temporal reasoning*. Then, the utilization of temporal probabilistic reasoning is required at the lower level of decision.

The use of a probabilistic temporal reasoning mechanism, besides the slow failures mentioned above, will also help to focus on the dynamic characteristics of the models in a power plant. Since the process of power generation has different phases, (e.g., start up, synchronization, steady state, and stop) different probabilistic models are required. For example, during the startup phase, the velocity of the turbine is the variable that will be substituted by the Megawatts generated during other phases. A mechanism that allows changes in the probabilistic model is required.

In addition to the two levels of decision, a new level of reasoning is required to detect when the fault is in the process, and not in the readings of the instruments. For example, the sensor validator may detect an erroneous reading from the turbine velocity given the temperature and pressure measures. However, it may be the case that there is a serious mechanical problem with the generator which may cause that the real velocity to go to a very low value.

Finally, the last step in the project will be the construction of a prototype which performs in a thermoelectrical power plant. This prototype requires a real time response. For this reason, different mechanisms of scheduling have to be developed, e.g., *any time* algorithms [Ibargüengoytia et al., 1995].

### Acknowledgments

Special thanks to Eduardo Morales who provided invaluable advice in the design of this methodology. Thanks also to the anonymous referees for their comments which improved this article. This research is supported by a grant from CONACYT and IIE under the In-House IIE/SALFORD/CONACYT doctoral programme.